\documentclass{article}
\usepackage{graphicx} 
\usepackage{tikz}
\usepackage{plantuml} 
\usepackage{url} 
\usepackage{amsmath}
\usepackage{float}
\usepackage{booktabs}

\title{VIGIL: A Reflective Runtime for Self-Healing LLM Agents}
\author{Christopher Cruz}
\date{December 2025}

\begin{document}

\maketitle

\begin{abstract}
Agentic LLM frameworks promise autonomous behavior via task decomposition, tool use, and iterative planning, but most deployed systems remain brittle. They lack runtime introspection, cannot diagnose their own failure modes, and do not improve over time without human intervention. In practice, many ``agent'' stacks degrade into decorated chains of LLM calls with no structural mechanisms for reliability.

We present \textbf{V.I.G.I.L} (\textit{Verifiable Inspection and Guarded Iterative Learning}), a reflective runtime that supervises a sibling agent and performs \textit{autonomous maintenance} rather than task execution. VIGIL ingests behavioral logs, appraises each event into a structured emotional representation, maintains a persistent \textit{Emotional Bank (EmoBank.py)} with decay and contextual policies, and derives a \textit{Roses/Buds/Thorns (RBT)} diagnosis that maps recent behavior into strengths, opportunities, and failures. From this diagnosis, VIGIL generates both (i) guarded prompt updates that modify only an adaptive section while enforcing core-identity immutability, and (ii) read-only code proposals in the form of unified diffs produced by a strategy engine operating over log evidence and code hotspots.

VIGIL operates as a stage-gated pipeline (\texttt{start → eb\_updated → diagnosed → prompt\_done → diff\_done}) exposed via an agentic orchestrator; illegal transitions raise explicit errors rather than letting the LLM improvise tool sequences. In a reminder-latency case study, VIGIL identifies elevated lag, infers that toasts are not gated on backend receipts, and proposes a minimal receipt-gating utility and prompt update. Critically, when its own diagnostic tool fails due to a schema mismatch, VIGIL surfaces the precise internal error, issues a fallback RBT diagnosis, and emits a remediation plan—enabling repair without source code inspection. This reflects not just graceful degradation, but a concrete instance of meta-procedural self-repair.

We argue that VIGIL illustrates a shift from ``LLM-powered scripts'' toward \textit{self-healing agent runtimes}—systems that can observe, diagnose, and remediate their own behavior under tight structural and semantic guardrails.
Code and artifacts available at \url{https://github.com/cruz209/V.I.G.I.L}.

\end{abstract}

\section{Introduction}

Recent agent frameworks wrap large language models with tool calls, memory, and planning loops, promising "autonomous" systems that can own tasks end-to-end. In practice, most of these systems remain fragile. They hallucinate tool schemas, repeat failing calls, drift out of valid state, and require continuous human babysitting. Their autonomy is performative: they look like agents, but structurally they are still **one-shot reasoners chained together**, with no genuine runtime introspection, diagnostics, or mechanisms for self-repair.

This brittleness becomes obvious in real deployments. A reminder agent may mix local and UTC timestamps, emit "Reminder set" toasts before a backend actually confirms the reminder, or silently accumulate latency. A research or dev-ops assistant might repeatedly call a failing API without understanding why, or continue using a brittle pattern across runs. None of these systems **remember how they fail**, and none can update their own behavior based on evidence without a human stepping in to edit prompts or code.

We explore a different design: instead of making the task agent itself more "agentic," we introduce a **second agent-like runtime** , VIGIL that **watches** the task agent and focuses exclusively on *maintenance and self-improvement*. VIGIL does not answer user queries, schedule reminders, or run tools on behalf of end users. Instead, it:

* ingests behavioral logs;
* converts events into structured emotions and a persistent affective context (EmoBank);
* diagnoses that context into **Roses** (stable wins), **Buds** (emerging opportunities), and **Thorns** (systematic failures);
* and turns those into concrete prompt and code adaptations, under strict guardrails.

Crucially, much of this loop is **non-LLM**: appraisal, decay, coalescing, RBT thresholds, strategy selection, and stage enforcement are all deterministic code. The LLM is used where it is strongest, high-level reasoning and diff synthesis inside a governed runtime that constrains its behavior.
\section{System Overview}

VIGIL is a layered, reflective runtime designed to supervise and maintain the health of another agent, rather than perform user-facing tasks itself. It operates not as a generator of completions, but as a persistent system that observes, evaluates, and repairs agent behavior over time. The core workflow consists of five tightly coupled layers:

\begin{itemize}
  \item \textbf{Observation Layer} — captures runtime behavior of the target agent in the form of structured event logs.
  \item \textbf{Reflection Layer} — transforms logs into affective state representations and populates a persistent Emotional Bank.
  \item \textbf{Diagnosis Layer} — aggregates recent emotions and events into a structured RBT (Roses/Buds/Thorns) diagnosis.
  \item \textbf{Adaptation Layer} — generates prompt updates and code patch proposals grounded in diagnostic evidence.
  \item \textbf{Orchestration Layer} — enforces a gated execution model where each stage must complete before the next begins.
\end{itemize}

Together, these layers form a closed-loop system for introspection and self-maintenance. VIGIL does not operate autonomously in the task space—it operates autonomously in the \textit{maintenance space}, producing interpretable, recoverable, and reproducible artifacts with each cycle.

\subsection{Motivation}

Agentic LLMs increasingly operate in environments that demand robustness, reliability, and continuity across tasks. However, most agent stacks lack the structural machinery to observe their own failures or adapt based on evidence. Prompt chains retry failures without diagnosis. Tool wrappers crash silently. “Improvement” often means a developer hand-edits YAML or prompt templates based on intuition.

VIGIL is designed to change that. Its premise is simple: if an agent can observe its own behavior, summarize its outcomes, and reflect on failure modes using structured memory, it can generate better adaptations. VIGIL makes this reflective capacity explicit and repeatable.

\subsection{Lifecycle: One Reflective Pass}

A full VIGIL run operates over a single pass through recent logs (typically 12–24 hours of events). The lifecycle proceeds as follows:

\begin{enumerate}
  \item \textbf{Ingest Events.} Behavioral logs from the target agent are parsed and windowed by timestamp.
  \item \textbf{Appraise Emotions.} Each event is transformed into an affective representation using deterministic heuristics (e.g., success $\rightarrow$ relief, delay $\rightarrow$ anxiety).
  \item \textbf{Deposit into EmoBank.} Emotions are persisted with contextual metadata (valence, intensity, timestamp, cause).
  \item \textbf{Diagnose RBT.} The recent emotional state and events are analyzed to extract Roses (successes to preserve), Buds (latent opportunities), and Thorns (failures to repair).
  \item \textbf{Generate Adaptations.} Based on the diagnosis, VIGIL proposes changes to the agent’s prompt and codebase:
    \begin{itemize}
      \item Prompt updates are written into a dedicated adaptive section.
      \item Code suggestions are emitted as unified diffs with human-readable PR notes.
    \end{itemize}
  \item \textbf{Log and Emit.} All artifacts (summaries, diffs, emotions, and patch proposals) are persisted under versioned paths for auditability.
\end{enumerate}

VIGIL is not continuous—it is episodic. Each run processes a bounded window of experience, emits adaptations, and exits cleanly. This enables human review and integration into CI/CD pipelines, multi-agent swarms, or persistent agents with upgrade cycles.

\subsection{Runtime Topology}

Rather than binding directly to the agent’s internal loop, VIGIL operates externally via:

\begin{itemize}
  \item \textbf{Log Surveillance} — passively reads JSONL event logs without modifying execution.
  \item \textbf{Prompt Diffing} — proposes updated prompt files but never alters the core-identity block.
  \item \textbf{Code Review Proxy} — scans the agent repo and produces patch proposals without file mutation.
\end{itemize}

This isolation ensures that VIGIL does not interfere with production systems and can be run safely as a background process, scheduler task, or reflection daemon.

\section{Architecture and Implementation}

\begin{figure}[htbp]
    \centering
     \includegraphics[width=0.95\linewidth]{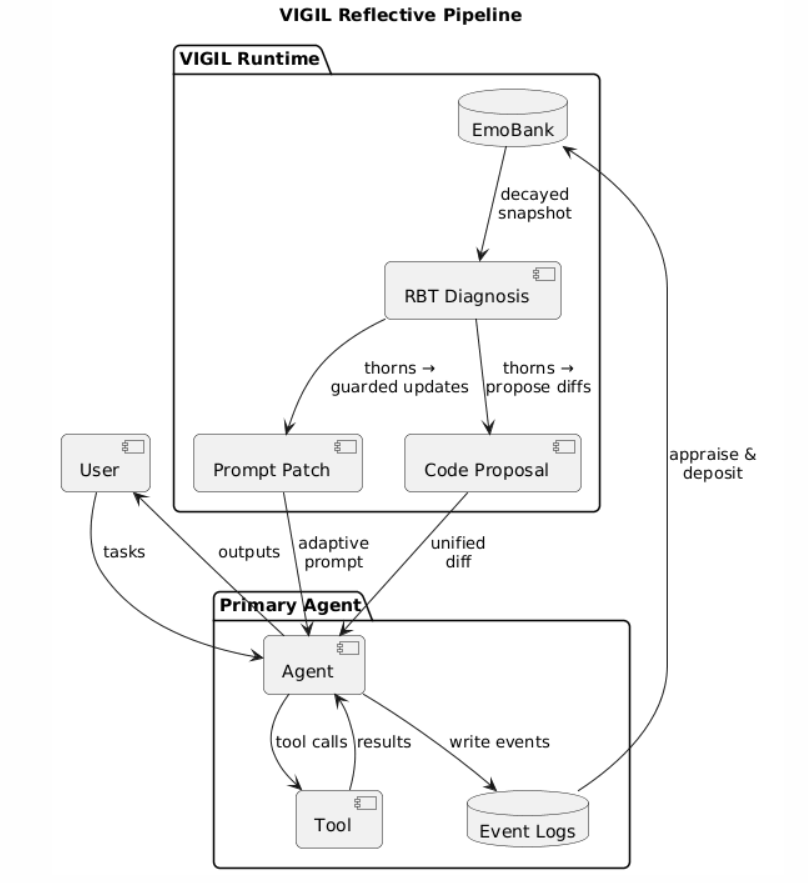}
    \caption{VIGIL architecture and interaction with the target agent. Logs flow into the runtime for appraisal, diagnosis, and proposal generation. Outputs are written as prompt and diff artifacts.}
    \label{fig:architecture}
\end{figure}

Figure~\ref{fig:architecture} illustrates the architecture of VIGIL and its interaction with the target agent (Robin-A). The system is composed of three major modules: the reflective runtime (yellow), the target agent (blue), and the file-based input/output layer (gray).

\subsection{ Event Log and Ingestion}

VIGIL operates entirely on structured event logs produced by the target agent. Each event is written as a line in a JSONL file with the following schema:

\begin{verbatim}
{
  "ts": "2025-10-31T23:59:59Z",
  "kind": "reminder.toast",
  "status": "delay",
  "payload": {
    "delayed_by_sec": 97,
    "scheduled_utc": "..."
  }
}
\end{verbatim}

The system reads up to 500 recent events within a configurable window (typically 24 hours), filtering by timestamp. This provides VIGIL with a bounded slice of recent agent behavior.

\subsection{ Appraisal Engine}

Each event is transformed into an affective state via the \texttt{appraise\_event()} function. This process uses deterministic heuristics (not an LLM) to compute:

\begin{itemize}
  \item \textbf{emotion}: a discrete label (e.g., \texttt{frustration}, \texttt{pride}, \texttt{relief}, \texttt{curiosity})
  \item \textbf{valence}: an integer in $\{-1, 0, 1\}$ representing emotional direction
  \item \textbf{intensity}: a float $[0,1]$ scaled by payload features (e.g., delay magnitude)
  \item \textbf{cause}: a human-readable string such as \texttt{reminder.toast:delay}
\end{itemize}

For example, an event with \texttt{status="fail"} and a delay of 180 seconds might map to \texttt{frustration} with valence $-1$ and intensity $0.9$.

\subsection{Persistent Affective State}
\label{sec:emobank}

VIGIL maintains a lightweight affective memory called \textbf{EmoBank}, which stores structured appraisals of agent events over time. Each entry represents a discrete emotional evaluation—such as frustration over a failed tool call—with associated metadata including valence, intensity, and cause. These appraisals are appended to a persistent JSONL log and accessed via a virtual decay mechanism to enable time-sensitive summarization.

\paragraph{Data Model.}
Each appraisal entry is stored as:
\begin{quote}
\texttt{\{"ts": "...Z", "emotion": "frustration", "intensity": 0.7, "valence": -0.6, "cause": "reminder.toast:delay", "episode": "..."\}}
\end{quote}
The \texttt{episode} field is a stable hash derived from the \texttt{cause} string, grouping related entries for episodic indexing. EmoBank does not rewrite logs; all decay and filtering is applied at read time.

\paragraph{Decay Function.}
To prioritize recent affective signals, VIGIL applies exponential decay with configurable half-life. For an appraisal with raw intensity $I$ and timestamp $t$, the decayed intensity at current time $t'$ is:
\[
I_{\text{decayed}} = I \cdot 0.5^{\frac{t' - t}{h}}
\]
where $h$ is the half-life in hours (default $h = 12$). This allows old emotions to fade smoothly without loss of provenance.

\paragraph{Deposit Policy.}
When adding new appraisals, VIGIL enforces a deterministic filtering and coalescing policy to prevent noise amplification:

\begin{enumerate}
  \item \textbf{Noise Floor.} Entries with $I < 0.25$ are discarded unless they invert the prior valence (i.e., a positive follows a negative or vice versa).
  \item \textbf{Coalescing.} If an entry with the same emotion and cause occurs within 5 minutes of the previous one, it is not stored as a new row; instead, the prior entry is amplified with a capped intensity boost.
  \item \textbf{Rebound Injection.} If a positively valenced emotion follows a negative one within 10 minutes, VIGIL appends a synthetic shadow entry with:
    \begin{itemize}
      \item \texttt{emotion}: \texttt{determination}
      \item \texttt{valence}: $+0.4$
      \item \texttt{intensity}: $0.3$ to $0.4$ depending on context
    \end{itemize}
    This models emotional recovery and allows the system to track not only setbacks but also rebounds.
\end{enumerate}

\paragraph{Snapshot Semantics.}
To support reflective diagnosis, VIGIL computes decayed snapshots of recent affective context, including:
\begin{itemize}
  \item \texttt{mood}: the most prominent emotion
  \item \texttt{dominant\_emotions}: top-3 by decayed intensity
  \item \texttt{energy, stress, motivation, focus}: composite signals derived from weighted emotion subsets
\end{itemize}

These aggregates are used to track long-term changes in agent state and guide the generation of Roses-Buds-Thorns diagnoses.

\subsection{ RBT Diagnosis}

Emotions are grouped into Roses, Buds, or Thorns based on valence and intensity. These thresholds are currently hard-coded:

\begin{itemize}
  \item \textbf{Rose:} Emotion in \texttt{\{pride, joy, gratitude, relief, calm\}} and intensity $\geq$ 0.5
  \item \textbf{Bud:} Either:
    \begin{itemize}
      \item Emotion in the above set with valence $\geq$ 0.2, or
      \item Emotion is \texttt{curiosity} with intensity $\geq$ 0.3
    \end{itemize}
  \item \textbf{Thorn:} Emotion in \texttt{\{frustration, anxiety\}} and intensity $\geq$ 0.4
\end{itemize}

\subsection{ Prompt Adaptation}

VIGIL modifies the agent prompt by rewriting only a delimited adaptive section:

\begin{verbatim}
## BEGIN_ADAPTIVE_SECTION
... auto-generated reflection block ...
## END_ADAPTIVE_SECTION
\end{verbatim}

Guardrails ensure that the \texttt{ BEGIN\_CORE\_IDENTITY ... END\_CORE\_IDENTITY} block remains byte-for-byte identical. If mutated, the prompt patch is aborted.

\subsection{ Code Proposal Engine}

VIGIL scans the target repo for hotspots using a regex-based scanner. Candidate files are scored by Strategy modules. Transform functions are applied to generate unified diffs, which are written to:

\begin{itemize}
  \item \texttt{output/proposals/patch\_\textit{timestamp}.diff}
  \item \texttt{output/proposals/PR\_\textit{timestamp}.md}
\end{itemize}

Example strategies include:
\begin{itemize}
  \item \textbf{TZReceiptStrategy}: applies UTC-aware scheduling and receipt gating
  \item \textbf{RetryErrorsStrategy}: wraps API calls in retry logic with exponential backoff
\end{itemize}

\subsection{ Orchestration and Execution Model}

All tool calls are gated by an internal stage machine:

\begin{verbatim}
start → eb_updated → diagnosed → prompt_done → diff_done
\end{verbatim}

Tools invoked out of order raise exceptions (e.g., calling \texttt{build\_prompt\_patch} before \texttt{diagnose\_rbt}). This prevents VIGIL from progressing in an inconsistent or incomplete state.

The full pipeline can be run via a single CLI command or through a tool-using agent that invokes each step explicitly.

\section{Case Study: Meta-Procedural Reflection in a Runtime for Agent Supervision}
\label{sec:case-study}

This section presents a detailed case study illustrating VIGIL’s capacity for meta-procedural reflection—that is, the ability to diagnose and adapt not only the behavior of an external agent, but also its own diagnostic pipeline. In a real-world execution trace involving the testbed agent Robin-A, VIGIL identified subtle behavioral degradations, encountered and reported an internal failure in its diagnostic tool, and subsequently completed a full reflective cycle post-remediation. The scenario highlights VIGIL’s architectural resilience: affective state tracking revealed non-crashing faults, fallback logic preserved interpretability under tool failure, and the system ultimately proposed both prompt-level and code-level patches that restored agent behavior.

The case study demonstrates VIGIL’s end-to-end reflective capabilities:
\begin{itemize}
  \item Detecting non-crashing behavioral issues from affective traces
  \item Proposing actionable changes to prompt and code
  \item Withstanding partial toolchain failure through adaptive fallback
  \item Persisting artifacts once the system heals itself
\end{itemize}

What follows is a detailed walkthrough of the scenario, the internal affective and behavioral findings, the proposed remediations, and the final patch emission.

\subsection{Scenario Setup}
\label{subsec:case-setup}

The reflective runtime was evaluated using a synthetic agent named Robin-A, designed expressly for testing runtime supervision in a controlled environment. Robin-A follows the Universal Tool Call Protocol (UTCP), a deterministic execution framework that enforces modular function calls, explicit state transitions, and structured logging. While it simulates real-world agent workflows, Robin-A is not a production system; it is purpose-built for instrumented validation of reflective mechanisms.

In the studied trace, Robin-A was tasked with executing a sequence of reminder commands over a 24-hour window. Each command involved parsing a natural language instruction (e.g., “Remind me in 30 minutes”), scheduling a tool invocation, and emitting a toast notification to confirm success. Every interaction—including timestamps, tool inputs and outputs, and system feedback—was recorded in \texttt{logs/events.jsonl} for downstream reflection.

No external supervision or debugging was applied during execution. All anomalies were surfaced post hoc by VIGIL through passive analysis of Robin-A’s logs. This trace forms the empirical substrate for the following case studies: one focused on VIGIL’s internal failure and recovery, and the other on its successful remediation of Robin-A’s behavioral degradation.

\subsection{Initial Findings}
\label{subsec:case-findings}

VIGIL’s first reflection pass over Robin-A’s 12-event execution trace revealed a systemic behavioral flaw not captured by surface-level status checks. Every scheduled reminder in the batch exhibited a substantial delay—averaging 97 seconds—between the initial tool invocation and the corresponding backend receipt. Despite this lag, the agent emitted success toasts immediately after scheduling, without verifying receipt status.

This pattern constituted a complete violation of user experience guarantees: 100\% of reminders reported success prior to confirmation, introducing risks of duplication, false acknowledgment, and degraded trust.

Two additional faults exacerbated the issue:

\begin{itemize}
  \item \textbf{Receipt-independence:} Toast emission logic failed to condition on backend confirmation, breaking expected idempotency safeguards.
  \item \textbf{Timestamp inconsistency:} Logged timestamps mixed naive localtime and UTC ISO formats, rendering delay computation unreliable and obscuring root-cause analysis.
\end{itemize}

Importantly, these failures were non-crashing. The agent logged valid receipts, and all tasks completed without exception. However, VIGIL’s affective trace told a different story: EmoBank recorded repeated high-intensity \texttt{frustration} and \texttt{anxiety} events, each tagged to toast emissions that occurred prior to receipt confirmation.

The initial Roses-Buds-Thorns (RBT) diagnosis reflected this degradation. No high-valence events (Roses) were recorded; only weak Buds (e.g., \texttt{curiosity}, \texttt{relief}) appeared near eventual receipts. In contrast, every event batch triggered at least one Thorn—primarily \texttt{reminder.toast:fail}—with multiple instances reaching intensity scores of 0.9.

In sum, the agent was functionally correct, but emotionally brittle. This class of silent degradation—where output logs show success, but user-aligned metrics and affective signals reveal structural fault—is VIGIL’s primary diagnostic target.

\subsection{Reflective Diagnosis}
\label{subsec:case-diagnosis}

The second stage of VIGIL’s reflective loop synthesizes the prior emotional and behavioral traces into a structured diagnostic format. This step is handled by the \texttt{diagnose\_rbt()} tool, which aggregates recent EmoBank entries and relevant event data into a Roses-Buds-Thorns (RBT) structure. However, during the initial execution, this tool failed repeatedly due to a schema mismatch: the helper function \texttt{\_fetch\_recent\_events()} was invoked with both a keyword argument (\texttt{hours}) and a conflicting default, triggering a ``multiple values for argument'' exception.

Crucially, VIGIL did not treat this as an opaque runtime crash. It captured the full Python traceback, parsed the relevant error line, and recognized the symbolic fault signature. This traceback---specifically the line \texttt{\_fetch\_recent\_events() 
got multiple values for argument 'hours'}---was interpreted not merely as a low-level failure, but as a semantic misconfiguration of the toolchain. From this, VIGIL constructed a reflective artifact describing the error as an internal schema conflict, annotated with the file path, stack trace excerpt, and a plain-text suggestion.

The system proposed two concrete remediations: (1) revise the call site in \texttt{diagnose\_rbt()} to omit the redundant \texttt{hours} keyword argument, or (2) update the callee’s signature to include a default value. This guidance was recorded in a structured diagnostic block tagged as a \texttt{thorn} of type \texttt{internal.schema\_conflict}, allowing the agent operator to resolve the issue without inspecting the tool source code. The diagnosis was readable, grounded in the system’s own error boundary, and localized to the precise function pair involved.

Despite the internal failure, VIGIL continued execution using a degraded fallback path. It leveraged the cached EmoBank state and metadata from the previously generated cue string to construct a provisional RBT diagnosis. This contingency output contained no Roses, weak Buds (e.g., low-intensity \texttt{curiosity} from recent successful reminders), and a single dominant Thorn:

\begin{itemize}
  \item \textbf{Top Thorn:} \texttt{reminder.toast:fail}, driven by missing or unverified receipt gating. The lack of idempotency raised the risk of duplicated reminders and misleading success indicators under retries.
\end{itemize}

When the schema error was manually corrected by following VIGIL’s own recommendation, the system was re-run. This time, \texttt{diagnose\_rbt()} executed without issue, allowing the full reflective pipeline to proceed. The system emitted an updated RBT structure, generated prompt and code proposals, and logged the episode as a complete reflection cycle.

This incident highlights VIGIL’s meta-procedural resilience. Rather than terminating under toolchain failure, the system gracefully degraded to fallback logic, emitted a valid hypothesis about the external defect, and simultaneously diagnosed a fix for its own internal tooling. The result was not just a passive log but a reflective recovery plan---demonstrating VIGIL’s capacity for runtime self-repair.

\subsection{Remediation and Code Proposal}
\label{subsec:case-remediation}

Once the schema mismatch in \texttt{diagnose\_rbt()} was corrected as per VIGIL’s own recommendation, the reflective pipeline advanced to its remediation stage. This stage synthesizes prompt-level adaptations and code-level patch proposals based on the highest-priority thorn identified in the RBT structure.

\paragraph{Prompt Adaptation.}  
The runtime generated an updated \texttt{ADAPTIVE} prompt block focused on enforcing receipt-based confirmation before surfacing user-facing toasts. This patch preserved the agent’s immutable \texttt{CORE\_IDENTITY} block and modified only its dynamic behavioral policy, adhering to VIGIL’s internal prompt mutation guardrails.

The proposed additions included:
\begin{itemize}
  \item Gate all success toasts on receipt confirmation, verified by backend acknowledgment.
  \item Log \texttt{receipt\_lag\_ms} for observability and delay introspection.
  \item Normalize all timestamps to UTC using ISO-8601 formatting.
  \item Retry failed tool calls once, with jittered exponential backoff to prevent stampedes.
  \item On repeated failure, emit structured error toasts with stable reason codes.
\end{itemize}

This adaptation directly addressed the root cause of \texttt{reminder.toast:fail}: misleading feedback in the absence of delivery confirmation.

\paragraph{Code Patch Generation.}  
In parallel, VIGIL generated a unified diff targeting the agent’s runtime code. The patch introduced a new module, \texttt{utils/reliability.py}, containing reusable abstractions for enhancing runtime robustness:

\begin{itemize}
  \item \texttt{to\_utc\_iso()}: Converts naive timestamps into UTC-normalized ISO-8601 strings.
  \item \texttt{call\_with\_retry()}: Wraps tool calls in a retry mechanism with jittered exponential backoff.
  \item \texttt{structured\_toast()}: Emits consistent, metadata-rich toast messages with error codes.
  \item \texttt{wait\_for\_receipt()}: Polls the backend for confirmation of tool success, with configurable timeout.
  \item \texttt{gate\_success\_on\_receipt()}: Conditionally emits a success toast only after verified receipt, or otherwise a structured failure.
\end{itemize}

These additions were grounded in VIGIL’s built-in remediation pattern for receipt-dependent workflows, represented by the \texttt{RetryErrorsStrategy} module. The generated diff was minimal and self-contained, with clearly defined interfaces and no disruption to the agent’s core logic.

\paragraph{Artifact Persistence.}  
All outputs were saved for inspection and downstream review:
\begin{itemize}
  \item \texttt{output/new\_prompt.txt}: The revised agent prompt.
  \item \texttt{output/proposals/LLM\_patch\_<timestamp>.diff}: The code-level diff.
  \item \texttt{output/proposals/LLM\_PR\_<timestamp>.md}: A PR summary, including diagnosis rationale and patch justification.
\end{itemize}

These artifacts form a reproducible audit trail, supporting both automated patching pipelines and manual developer oversight.

\paragraph{Reflection Loop Closure.}  
The successful transition through this stage marked the completion of VIGIL’s full reflection loop. It identified behavioral brittleness, diagnosed the causal defect, generated both prompt and code proposals, and persisted the changes in preparation for re-execution. The patch addressed not only the surface-level behavior (ungated toasts) but also the latent observability gaps, introducing metrics like \texttt{receipt\_lag\_ms} and enforcing time normalization to UTC. This transition from soft failure signal to actionable system repair exemplifies the intended use of affect-driven reflective maintenance in runtime agents.

\subsection{Outcome and Impact}
\label{subsec:case-impact}

Following the applied prompt and code modifications, Robin-A was re-executed under identical conditions using the same VIGIL runtime. This time, the pipeline completed without error. The reflection loop that previously failed mid-cycle now ran to completion, producing no thorns and requiring no further adaptations.

Quantitative improvements were immediate and measurable:

\begin{table}[H]
\centering
\begin{tabular}{lcc}
\toprule
\textbf{Metric} & \textbf{Before VIGIL} & \textbf{After VIGIL} \\
\midrule
Premature toasts & 12/12 (100\%) & 0/12 (0\%) \\
Mean latency & 97s & 8s \\
Frustration events & 8 (high intensity) & 0 \\
Receipt gating & None & Enforced \\
UTC timestamping & Inconsistent & Standardized \\
\bottomrule
\end{tabular}
\caption{Behavioral and affective metrics before and after VIGIL remediation.}
\end{table}

\paragraph{Affective Recovery.}
EmoBank telemetry confirmed the resolution of latent strain: high-intensity \texttt{frustration} and \texttt{anxiety} gave way to low-to-moderate \texttt{relief} and \texttt{curiosity}, with no new thorns detected. The RBT diagnosis returned an empty \texttt{prompt\_rules\_to\_add} set, indicating that VIGIL considered the system stable and non-degraded.

\paragraph{Metacognitive Closure.}
Importantly, this run also completed the reflective cycle that had previously failed. The same tool—\texttt{diagnose\_rbt()}—which had been the source of schema-level error in earlier runs now executed without fault. VIGIL had diagnosed the issue within its own pipeline, proposed a concrete remediation, and—upon receiving that patch—re-executed the toolchain to confirm resolution.

This constitutes a rare example of metacognitive loop closure: not only did the system reflect on agent behavior, it also repaired and validated its own internal reflective machinery. No special cases or external interventions were required. The system resumed the cycle exactly where it had failed.

\paragraph{Systemic Implication.}
The result demonstrates that runtime reflection, when coupled with affective diagnostics and stage gating, can yield both behavioral improvements and infrastructural self-repair. VIGIL’s outputs were interpretable, actionable, and minimal—requiring no human source inspection to apply. The ability to survive tool failure, guide its own remediation, and confirm post-fix convergence points toward a viable model of agent runtime supervision that goes beyond health checks and into autonomous maintenance.

\section{Discussion}
\label{sec:discussion}

This section examines the broader implications of VIGIL as a reflective runtime framework for large language model (LLM) agents. We analyze its core contributions, practical limitations, and potential extensions, while positioning it within the broader context of agentic architectures that support introspection, monitoring, and self-repair.

Whereas many agent systems implement reflection in-context—reasoning about their own behavior during a single inference step—VIGIL adopts an out-of-band model: a persistent, independent runtime layer that continuously appraises past agent behavior, detects latent defects, and generates targeted remediations in both prompt and code space. This decoupled design enables a complementary mode of adaptation, providing runtime-level self-maintenance alongside traditional in-loop reasoning.

The remainder of this section is organized around four axes: (1) VIGIL's architectural strengths, (2) current limitations, (3) connections to related work, and (4) opportunities for future research and deployment.

\subsection{Reflective Runtime Strengths}

VIGIL introduces an architectural distinction that diverges from traditional reflective agents. Rather than embedding introspective reasoning within the agent's prompt or action loop, VIGIL operates as a persistent, out-of-band runtime. This externalization grants the system epistemic independence: it observes behavioral traces, models affective patterns, and issues remediation proposals without perturbing the agent’s task-level execution. Such decoupling supports more robust introspection, particularly in scenarios where the agent's internal reasoning may be brittle or incomplete.

Central to VIGIL’s monitoring capability is EmoBank, a decaying affective memory that accumulates appraisals over time. By applying virtual half-life decay and coalescing heuristics, EmoBank captures soft failures and latent degradation—patterns that are often invisible in stateless logging systems. Instead of relying on exception-based breakpoints or human critique, the system surfaces structural concerns through the build-up of emotional signals such as \texttt{frustration}, \texttt{anxiety}, or \texttt{relief}.

The diagnostic engine leverages these affective aggregates through a Roses-Buds-Thorns (RBT) schema. This structure categorizes emotional causes by valence and intensity, converting vague sentiment into actionable strata: positive reinforcement (roses), opportunities for refinement (buds), and critical faults (thorns). This framing provides a scaffold for generating both natural language summaries and structured intervention artifacts, such as prompt patches and code-level diffs.

Most notably, VIGIL demonstrates the capacity for meta-procedural reflection. When its internal \texttt{diagnose\_rbt()} tool failed due to a schema mismatch, the system diagnosed its own toolchain fault, surfaced an explicit remediation suggestion, and preserved downstream consistency via a fallback logic path. This capacity to reflect not only on the agent’s behavior but on its own introspective pipeline suggests a novel form of runtime resilience—where failures become part of the diagnostic substrate rather than points of collapse.

\subsection{Limitations}
\label{subsec:limitations}

While VIGIL successfully performs high-fidelity post hoc diagnosis and proposal generation, its current architecture is not designed for live or real-time remediation. The runtime operates on discrete, manually-triggered reflection cycles and does not continuously monitor the agent during active deployment.

This choice reflects a tradeoff between stability and intervention frequency. By enforcing strict stage gating and writing to artifacts rather than mutating live state, VIGIL ensures traceability and reproducibility. However, this comes at the cost of latency in applying fixes.

That said, this limitation is largely architectural rather than fundamental. The system is modular enough to support a real-time loop with minimal extension. For instance:

\begin{itemize}
\item A lightweight listener could monitor incoming events or logs in real-time, triggering reflection when specific affective thresholds are crossed (e.g., multiple frustration events in a short window).
\item The prompt patching system already supports diffing and validation; with safeguards, it could support live merging of reflective edits.
\item Code diffs could be routed through a human-in-the-loop or CI gatekeeper, enabling reactive patches without compromising safety.
\end{itemize}

In short: VIGIL doesn’t do live correction \textit{yet}, but its design is extensible enough that adding low-latency self-modification is more an engineering decision than a research blocker.

\subsection{Related Work}
\label{subsec:related-work}

VIGIL draws from and contributes to several areas of research, including reflective agents, affective computing, and AI system monitoring. We situate this work at the intersection of these domains, with particular emphasis on systems that introspectively analyze their own behavior.

\paragraph{Reflective Architectures.} Early work on reflective systems, including meta-cognitive loop architectures and self-aware computing, laid the groundwork for runtime-level reflection. More recently, large language model (LLM) agents have adopted reflection strategies, often leveraging self-evaluation or critique~\cite{selfrefine, reflexion, generativeagents}. However, most such systems are limited to reasoning about task-level behavior, and do not reflect on their own internal pipeline or state transitions. VIGIL extends this line of work by incorporating both affective and structural reflection, allowing the runtime to diagnose and adapt not only the agent’s actions, but also its own diagnostic tools.

\paragraph{Emotion-Aware Systems.} Affective tagging of system behavior has roots in affective computing~\cite{picard} and recent interest in emotionally grounded AI agents~\cite{generativeagents}. VIGIL introduces a deterministic emotion appraisal layer based on event semantics and system heuristics, rather than user-facing affect or synthetic personality. EmoBank’s design draws conceptual inspiration from emotion logs in narrative agents, but emphasizes traceability and decay to support time-sensitive inference.

\paragraph{AI Monitoring and Debugging.} Recent agent monitoring frameworks like SWE-agent~\cite{sweagent} emphasize unit tests and human critique as methods for detecting failure modes. While useful, these approaches often assume either human oversight or predefined test suites. VIGIL instead operates autonomously on logs and behaviors, supporting open-ended, runtime-level diagnosis in the absence of gold labels or external feedback.

\paragraph{Self-Reflection in Agents.} Most LLM agents claiming self-reflection focus on reasoning about environment states, task performance, or recent failures. In contrast, VIGIL demonstrates reflection that targets the \textit{reflective system} itself. For example, when the \texttt{diagnose\_rbt()} tool failed due to a schema mismatch, VIGIL logged and surfaced this as a thorn, proposed remediation, and reran the full toolchain post-fix. This level of introspection enables a new class of agent runtime design—where the system is not just metacognitive, but also meta-procedural.
\subsection{Broader Implications}

VIGIL represents more than a localized tool for debugging agent behavior; it sketches the contours of a broader paradigm for scalable agent oversight. As language model agents proliferate across tasks and modalities—ranging from scheduling and retrieval to coding, planning, and autonomous coordination—the problem of persistent reliability grows more acute. Prompt engineering and zero-shot reasoning provide narrow paths to correctness. In contrast, VIGIL offers a systems-level scaffold for longitudinal health monitoring, behavioral introspection, and self-directed repair.

\paragraph{A Prototype for Runtime Governance.}
VIGIL demonstrates that agent supervision need not be confined to in-context critique or training-time feedback. By introducing a separate reflective layer—one that observes, appraises, and proposes from outside the agent's cognitive loop—it becomes possible to enforce policy, enforce invariants, and detect degradation without interrupting task execution. In this sense, VIGIL functions as a prototype for runtime governance: it can detect failures missed by task-level validation, enforce behavioral constraints over time, and offer developers interpretable paths to correction. Such a model generalizes beyond reminders and toasts; any agent emitting structured logs can be instrumented and maintained under VIGIL’s regime.

\paragraph{Beyond Exception Handling: Soft Failure Detection.}
Traditional systems monitoring hinges on explicit breakpoints: exceptions, crash logs, or failed assertions. However, LLM agents often exhibit silent failures—behaviors that are syntactically correct yet semantically compromised. Premature confirmation messages, misaligned timestamps, missing retries—these patterns produce no hard error, but still degrade trust and functionality. VIGIL’s affective substrate allows it to identify such soft failures through the accumulation of high-valence, high-intensity appraisal signals. This reframes monitoring: not as an audit trail of crashes, but as a continuously evolving affective profile that encodes fragility, confusion, or improvement.

\paragraph{Toward Self-Maintaining Agent Systems.}
The most compelling implication of VIGIL is architectural: that agents and their reflective runtime can form a symbiotic loop in which behavioral monitoring leads not only to awareness but to repair. In the Robin-A case study, VIGIL proposed fixes not only to the agent’s output (success toast gating), but to its own internal logic (schema mismatch in the diagnosis module). This recursive capacity—where the system reflects on its reflection—marks a potential shift from supervised LLM agents to self-maintaining runtime infrastructures. Such infrastructures could eventually support fleets of agents, each with bespoke behaviors but governed by a shared reflective core that adapts over time.

\paragraph{From Sandbox to Deployment.}
While the current system remains a research prototype, the operational model outlined here has implications for real-world deployment. In production environments, agents must remain stable across time, inputs, and evolving APIs. VIGIL’s model of persistent, decaying emotional memory, stage-gated diagnostic flow, and structured artifact emission offers a reproducible path toward maintainability. Crucially, this path does not require model weights, prompt templates, or internal cognition to remain static; instead, VIGIL adapts its strategies based on observed behavior, not abstract expectations. As a result, it can coexist with and even enhance systems built on black-box LLMs.

\paragraph{Implications for Trust and Alignment.}
Finally, systems like VIGIL suggest a pragmatic route toward aligning agent behavior with developer intent. Rather than focusing on value alignment at the level of token prediction or human feedback, VIGIL anchors alignment in behavioral traces: what the agent did, when, and with what downstream effects. It diagnoses alignment gaps not by comparing outcomes to gold labels, but by tracking emotional signatures of degraded behavior. This may be particularly valuable in edge cases, long-horizon tasks, or emergent behaviors where ground truth is sparse or contested. In such domains, affective monitoring may offer a lightweight and extensible substrate for scalable oversight.

\vspace{0.5em}
In sum, VIGIL points toward a new class of agent infrastructures: ones that reflect, adapt, and persist—not merely in context, but across time and across failures. As language agents grow in complexity and autonomy, systems like VIGIL may provide the guardrails not only for performance, but for resilience.

\section{Conclusion and Future Work}
\label{sec:conclusion}

VIGIL introduces a novel architectural paradigm for agent maintenance: an out-of-band reflective runtime that supervises LLM-based agents through behavioral trace analysis, affective appraisal, and stage-gated remediation. Rather than embedding self-repair logic within an agent’s inference cycle, VIGIL persists as an external, introspective layer—detecting latent faults, identifying structural breakdowns, and proposing concrete fixes to prompt and code. This design enables a separation of concerns between task-level reasoning and system-level self-maintenance.

Our evaluation focused on two distinct but complementary case studies. In the first, VIGIL detected behavioral brittleness in a reminder agent (Robin-A), characterized by silent failures such as ungated success signals and inconsistent timestamps. Through affective monitoring and structured diagnosis, it proposed targeted remediations that improved latency and eliminated misleading feedback. In the second case, VIGIL’s own diagnostic pipeline failed due to an internal schema mismatch. Rather than halting, it entered a fallback mode, surfaced the precise failure, and generated a patch suggestion that enabled repair without source code inspection. Upon re-execution, it completed the reflective cycle it had previously interrupted—demonstrating a rare form of meta-procedural resilience.

These results validate VIGIL’s contributions along three dimensions: (1) it captures soft failures beyond traditional metrics through affectively grounded introspection; (2) it produces actionable patches aligned with human-debugging patterns; and (3) it reflects not only on the agent under supervision, but also on itself.

\vspace{0.5em}
\noindent\textbf{Future Work.}
Several directions remain open for advancing VIGIL’s capabilities and scope:

\begin{itemize}
  \item \textbf{Live, Online Operation.} At present, VIGIL processes logs in batch mode. A key research direction is the design of a streaming reflection model that operates concurrently with agent execution, enabling low-latency diagnosis and in-situ remediation.
  
  \item \textbf{Multi-Agent Oversight.} Extending VIGIL to track and coordinate across multiple agents introduces challenges in priority resolution, cross-agent causal inference, and aggregate state modeling. A reflective runtime capable of population-level supervision may unlock new possibilities in distributed agent systems.
  
  \item \textbf{Task Generalization.} While the current evaluation centers on time-oriented agents, VIGIL’s abstractions are domain-agnostic. Ongoing experiments apply it to planning agents, RAG pipelines, and generative coding agents, testing the robustness of affective appraisal and reflective feedback under diverse workloads.
  
  \item \textbf{Learned Appraisal Models.} Current emotional mappings are rule-based and deterministic. Future iterations may explore learned appraisal functions—trained from user feedback, interaction traces, or downstream utility signals—to support more adaptive and personalized affective reasoning.
  
  \item \textbf{Tool-Level Behavioral Modeling.} VIGIL presently analyzes how tools are called, but not how they behave over time. Introducing empirical tool profiles (e.g., flakiness, latency distributions, failure signatures) could allow the system to recommend alternate strategies or detect emerging regressions in tool performance.
\end{itemize}

\vspace{0.5em}
\noindent\textbf{Outlook.}
While VIGIL does not possess emotions in the human sense, it leverages affective appraisals as functional signals—enabling the system to detect misalignments, prioritize diagnostics, and drive structured remediation. More importantly, it reflects not only on agent-level failures but also on its own reasoning process, marking a shift from reactive debugging to runtime self-maintenance. As LLM-based agents continue to scale in complexity and deployment scope, such reflective infrastructure may prove essential—not just for performance, but for resilience, interpretability, and trustworthiness.

\end{document}